\documentclass[10pt,twocolumn,letterpaper]{article}

\usepackage{cvpr} 
\usepackage{times}
\usepackage{epsfig}
\usepackage{graphicx}
\usepackage{amsmath}
\usepackage{amssymb}

\usepackage{cite}
\usepackage{booktabs}
\usepackage{array}
\usepackage{multirow}
\usepackage{subcaption}
\usepackage{xcolor}
\usepackage{url}

\usepackage{adjustbox}

\usepackage[most]{tcolorbox}

\usepackage[breaklinks=true,letterpaper=true,colorlinks,bookmarks=false]{hyperref}

\graphicspath{{figures/}}
\newcommand{\best}[1]{\textbf{#1}}
\newcommand{\ca}{$\downarrow$c/$\uparrow$a}
\newcommand{\ac}{$\downarrow$a/$\uparrow$c}

\begin{document}

\title{Improving Low-Resolution Face Recognition under Limited Data: How Synthetic Data Generation Can Close the Domain Gap}

\author{Luis S. Luevano \quad \"{U}nsal \"{O}zt\"{u}rk \quad
Hatef Otroshi Shahreza \quad
Anjith George \quad
S\'{e}bastien Marcel\\
Idiap Research Institute, Switzerland\\
{\tt\small \{luis.luevano, unsal.ozturk, hatef.otroshi, anjith.george, sebastien.marcel\}@idiap.ch}
}

\maketitle

\begin{abstract}

Face Recognition (FR) systems in surveillance settings often encounter \emph{Low Resolution} (LR) faces, those whose face region falls below the standard $112\times 112$ input size. While labelled High Resolution (HR) training data is abundant, labelled \emph{native}-LR data, and above all \emph{paired} native-LR/HR data, is scarce.
One workaround is to \emph{synthesize} LR data from the available HR faces, but how much synthesis effort is repaid in recognition accuracy remains unclear. We present a study of \emph{simple} synthetic generation strategies for a compact, edge device-oriented face recognition system, spanning interpolation-based degradation, knowledge distillation, a Prepended Domain Transformer (PDT), Real-ESRGAN-style degradation, and a learned Super Resolution (SR) front-end with an identity-aware loss. We evaluate these strategies on synthetic cross-resolution face benchmarks (LFW, CFP-FP, AgeDB-30) and on TinyFace, a \emph{real-world} native LR dataset, and expose a \emph{synthetic--real gap}: the degradation setting that is optimal on synthetic benchmarks is not the one that is optimal on real LR.
We find that more synthesis effort does not help monotonically: the learned SR front-end does not surpass a direct feed of the aligned LR image into a strong backbone, while simple interpolation augmentation of a compact backbone is the only synthesis that improves over its own baseline.
We conclude that generative methods for LR face recognition must be validated on real LR and against a direct-feed baseline, and release our pipeline at \url{https://idiap.ch/paper/synth-lrfr}.
\end{abstract}

\section{Introduction}
\begin{figure}[t]
  \centering
  \begin{subfigure}{0.45\linewidth}\includegraphics[width=\linewidth]{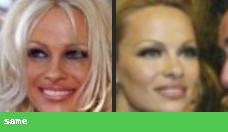}\caption*{\scriptsize$56\times56$\,px}\end{subfigure}\hfill
  \begin{subfigure}{0.45\linewidth}\includegraphics[width=\linewidth]{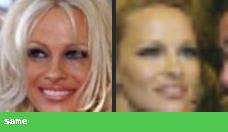}\caption*{\scriptsize$28\times28$\,px}\end{subfigure}\hfill
  \begin{subfigure}{0.45\linewidth}\includegraphics[width=\linewidth]{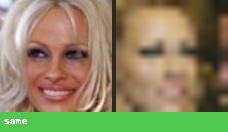}\caption*{\scriptsize$14\times14$\,px}\end{subfigure}\hfill
  \begin{subfigure}{0.45\linewidth}\includegraphics[width=\linewidth]{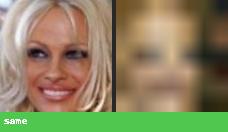}\caption*{\scriptsize$7\times7$\,px}\end{subfigure}
  \caption{A verification pair at decreasing target resolution, each upsampled
  back to $112$\,px. Identity cues vanish progressively.}
  \label{fig:ladder}
\end{figure}

Modern face recognition (FR) systems are highly accurate on High Resolution (HR) images in uncontrolled conditions, even on edge devices~\cite{martinez-diaz_benchmarking_fr_2021, martinez-diaz2019shufflefacenet, george2023edgeface,luevano2025swiftfaceformer}. Real-world deployments, however, frequently encounter Low Resolution (LR) faces as subjects can be captured at long distances and at different heights. After face detection and alignment, the usable facial region may span only a few dozen pixels (Fig.~\ref{fig:ladder}), and upsampling naively does not restore identity information~\cite{luevano2021lrsurvey}. More importantly, this LR setting suffers from \emph{limited data}, and the scarcity is
specific: labelled HR training data is plentiful, whereas labelled \emph{native}-LR
identities, and above all \emph{paired} native-LR/HR images of the same subject, are
costly to collect and largely unavailable. 
Beyond average accuracy, this setting also raises fairness concerns, since degradation can affect demographic groups unequally.
Synthesizing LR data and understanding effective degradation strategies for Low Resolution Face Recognition is therefore highly relevant to improve recognition performance under LR-side data scarcity.

In critical surveillance applications (e.g.\ watchlist screening, forensics), it is typical to match a high-quality enrollment (a cooperative capture, an identity document, or a mugshot) against a probe that may be only $16$--$32$ pixels across. This process therefore involves placing an LR probe into the same embedding space as an HR gallery template, which we call the HR$\to$LR setting. We also study the symmetric LR$\to$LR case (both images degraded) as a harder bound. 
The artifacts present when capturing probes at low resolutions include residual blur, sensor noise, and compression, forming a domain shift to which an HR-trained recognizer was never exposed. This is also why adapting the probe is often an employed strategy for LR FR.

A common approach synthesizes LR training data by downsampling HR faces and trains (or adapts) the recognizer on it. The synthesis is usually a fixed down-/up-sampling chain, and methods are validated on \emph{synthetically} degraded versions of standard verification sets because paired native-LR data is costly to collect. This poses the question: \emph{does a method tuned on synthetic LR actually work on real LR?}

We answer this for a compact, edge device-oriented backbone, EdgeFace~\cite{george2023edgeface}, and we use the answer to motivate and then evaluate generative alternatives. We compare five adaptation strategies (Sec.~\ref{sec:methods}) under the following protocol: interpolation augmentation, knowledge distillation, a native $32$\,px stem trained on Real-ESRGAN-style degradation, a Prepended Domain Transformer (PDT)~\cite{george2022pdt}, and a learned super-resolution front-end. To find the synthetic--real gap in performance, we evaluate everything on \emph{real} native-LR TinyFace under two alignment pipelines and compare it to synthetic LR FR benchmarks. 

Our contributions are the following:
\begin{itemize}
  \item We present a systematic study of synthetic data generation for LR face recognition on a compact, edge device-oriented backbone under limited LR data, under various generation efforts: deterministic interpolation, simulating Real-ESRGAN-style degradation, and a learned identity-aware super-resolution front-end.
  \item We provide an analysis on the \emph{synthetic--real domain gap} by evaluating the same models on synthetic cross-resolution benchmarks (LFW, CFP-FP, AgeDB-30) and on the real native-LR set TinyFace, isolating how the choice of degradation and target resolution governs transfer to genuinely captured LR faces.
  \item We release a pipeline and provide practical guidance for matching generation effort to the deployment setting (compact versus strong backbone), arguing that low-resolution face recognition should be validated on real LR and against a direct-feed strong-backbone baseline rather than on synthetic degradation alone.
  \item We complement the accuracy study with a demographic fairness assessment on RFW, and find that the average-accuracy gains from LR-aware synthesis do not translate into reduced demographic bias.
\end{itemize}
\section{Related Work}
\textbf{Efficient face recognition.} Margin-based losses such as CosFace~\cite{wang2018cosface} and ArcFace~\cite{deng2019arcface}, trained on web-scale data~\cite{zhu2021webface260m} with scalable classification heads~\cite{an2021partialfc}, define modern face recognition. A related line of work compresses these models for edge computing devices: EdgeFace~\cite{george2023edgeface} adapts the hybrid CNN--Transformer EdgeNeXt~\cite{maaz2022edgenext} with a low-rank linear layer, and lightweight designs were compared in the EFaR~2023 competition~\cite{kolf2023efar}. We adopt EdgeFace for benchmarking our synthesis approaches, as the LR FR problem is more pronounced when using such compact and deployable models, compared to their performance in HR FR.

\textbf{Low- and cross-resolution face recognition.} A large body of work targets
recognition when probes are degraded, most commonly by exposing the recognizer to
degraded data during training so that high- and low-resolution embeddings become
comparable~\cite{massoli2020crossres}. Approaches fall into three families:
resolution-robust \emph{training} (downsampling augmentation, cross-resolution
losses, feature adaptation)~\cite{massoli2020crossres}; \emph{distillation} of an
HR teacher into an LR student~\cite{ge2020lrkd,hinton2015distilling}; and
\emph{recovery}, in which a super-resolution or restoration module reconstructs
detail before recognition. The synthetic degradation used to build the training
and test data is usually treated as an implementation detail; we instead show
that this choice governs whether the learned robustness transfers to real LR faces, and we evaluate one method from each family on the same real
native-LR protocol.

\textbf{Super-resolution and blind face restoration for recognition.} Sub-pixel
convolution~\cite{shi2016espcn} and GAN-based super-resolution~\cite{wang2018esrgan}
recover high-frequency detail, and blind face restoration models such as
FaceMe~\cite{liu2025faceme} reconstruct plausible faces from severely degraded
inputs. Restoration can improve downstream recognition on low-quality
images~\cite{bengherabi2023boosting,martinezdiaz2023effectiveness}, but when
optimized for perceptual quality alone it can hallucinate identity-inconsistent
detail and even reduce accuracy. We therefore tie our super-resolution objective
to the recognizer embedding and keep the front-end light enough for an edge model.

\textbf{Generative degradation synthesis.} Real-ESRGAN~\cite{wang2021realesrgan}
models real-world degradation with a high-order random pipeline of blur,
resizing, noise, and compression. We repurpose this generator to
\emph{synthesize} realistic low-resolution training faces, a form of synthetic
data generation suited to our setting, where paired native-LR faces are
unavailable.

\textbf{Heterogeneous face recognition and domain transformers.} When low
resolution is treated as a separate modality, the recognizer can be frozen and
only the input adapted. The Prepended Domain Transformer
(PDT)~\cite{george2022pdt} and Domain Invariant Units~\cite{george2024dinu}
follow this recipe, attractive because the expensive backbone is never retrained.
We evaluate how far such input-side adaptation transfers to genuine native-LR data.

\section{Synthetic Data Generation and Adaptation Strategies}
\label{sec:methods}

We frame low-resolution adaptation as two coupled choices: how much effort to
spend \emph{synthesizing} the LR domain, and how to \emph{adapt} the recognizer
to it. We organize our methods along the following synthesis efforts:
\emph{low} is a fixed interpolation chain (downsample, then upsample back to
$112$\,px), with no learned or stochastic component; \emph{medium} is a
stochastic, high-order degradation model (Real-ESRGAN-style: blur, resizing,
noise, and compression) that mimics real sensor and optics degradation; and
\emph{high} adds a learned generative inverse, an identity-aware super-resolution
(SR) front-end that reconstructs an HR-like image. These data-side choices are
orthogonal to the recognizer-side adaptation (retraining the backbone, distilling
from a teacher, or freezing it and adapting only the probe), which we vary
independently; all models use the compact EdgeFace-S backbone unless a frozen,
stronger backbone is stated.

All synthetic data are derived from HR data, with each setting
denoted ``$s\,\downarrow\!d/\!\uparrow\!u$'' (downsample to $s$\,px with $d$,
upsample to $112$\,px with $u$; e.g.\ ``$28$\,\ca'' is cubic-down, area-up). For
training, interpolation yields one set per setting and the Real-ESRGAN pipeline
stores a native $32$\,px copy paired with its $112$\,px source. For evaluation, we
degrade the verification pairs of standard benchmarks in two modalities, HR$\to$LR
(one image degraded) and LR$\to$LR (both), and also test on the real native-LR set.

\subsection{Synthesis strategies}
\paragraph{Low effort: interpolation augmentation.}
The simplest adaptation downsamples WebFace4M~\cite{zhu2021webface260m} faces to
$s\in\{56,28,14,7\}$ with $\downarrow\in\{$area,cubic$\}$, upsamples back to
$112$\,px with $\uparrow\in\{$area,cubic$\}$, and retrains EdgeFace-S end-to-end
with CosFace~\cite{wang2018cosface} (PartialFC~\cite{an2021partialfc},
AdamW~\cite{loshchilov2019adamw}, $100$ epochs). 
This scheme does not make use of HR data, and considers LR to be purely augmentation.
Our comparison set is formed by the combinations of four reliably converging resolution/interpolation settings over $\{$area,cubic$\}{\times}\{$area,cubic$\}$.

\paragraph{Medium effort: generative degradation synthesis.}
To synthesize realistic LR training faces under limited native-LR data, we port
the Real-ESRGAN~\cite{wang2021realesrgan} high-order degradation generator. Per
image, a first-order block applies a random blur kernel (iso-/anisotropic Gaussian,
generalized Gaussian, plateau, or sinc), a random resize (factor $\in[0.15,1.5]$),
Gaussian/Poisson noise, and a JPEG cycle; a second-order block (probability
$0.8$) repeats blur/resize/noise; a final resize to $32\times32$ adds an optional
sinc low-pass and JPEG. We store the $32$\,px LR JPEGs alongside the original
$112$\,px HR. We also train a \emph{native $32$\,px variant}: EdgeFace-S with its $4\times4$
stride-4 stem replaced by a $1\times1$ convolution, taking the $32$\,px image directly without upsampling.

\begin{figure}[t]
  \centering
  \includegraphics[width=1.05\linewidth]{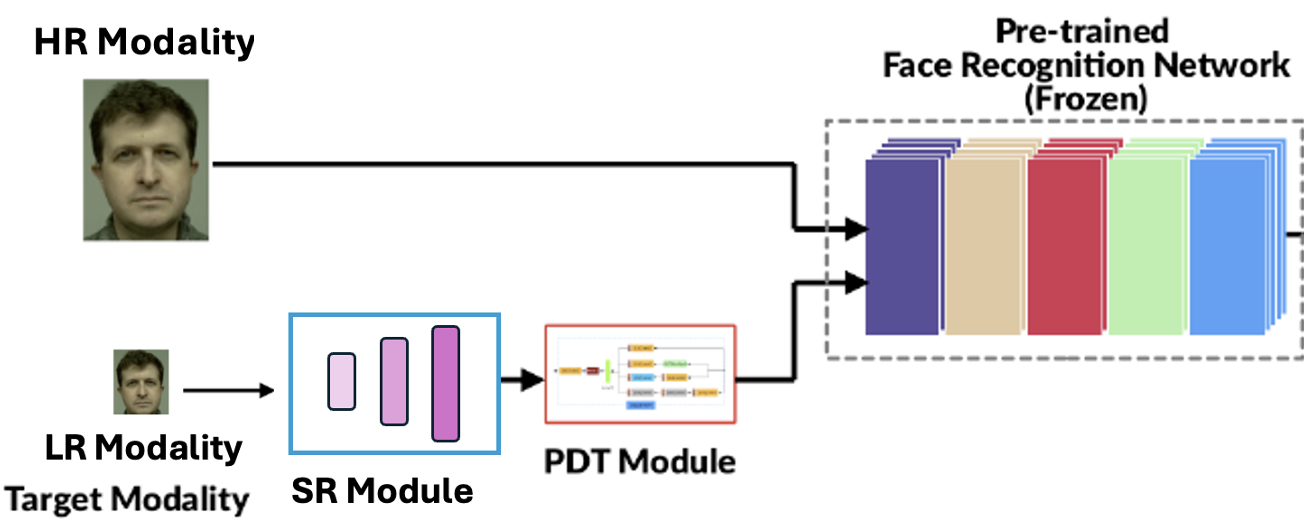}
  \caption{Input-side adaptation: a learned SR upsampler and a PDT translator map
  a $32$\,px native-LR face into the input domain of a \emph{frozen} backbone.
  Only the upsampler is pre-trained (identity-aware in Stage 1), then
  both upsampler and PDT module are trained with contrastive learning (Stage 2).}
  \label{fig:pipe}
\end{figure}

\paragraph{High effort: learned super-resolution front-end.}
We move upsampling \emph{inside} the model: the image stays at $32$\,px and the
upscale operator is learned. Two $\times2$ sub-pixel (PixelShuffle)
stages~\cite{shi2016espcn} take $32\!\to\!64\!\to\!128$\,px, then four
valid-padded $k{=}5$ convolutions trim $16$\,px ($128\!\to\!112$); no
interpolation function is used. Two feature-block variants of a plain
sub-pixel stack (ESPCN-style, $\approx$$0.86$\,M parameters) and an RRDB trunk
(ESRGAN/Real-ESRGAN-style~\cite{wang2018esrgan,wang2021realesrgan},
$\approx$$6.5$\,M) are studied. The upsampler is prepended to a PDT translator in front of a
\emph{frozen} backbone (EdgeFace-base, WebFace12M-pretrained), giving the path
\texttt{Upsampler}$\to$\texttt{PDT}$\to$frozen backbone (Fig.~\ref{fig:pipe}).
\emph{Stage 1 (SR pretraining)} regresses on (LR-$32$, HR-$112$) pairs with an
identity-aware perceptual term,
\begin{equation}
\begin{aligned}
\mathcal{L}_{\mathrm{SR}} = \ell_1\!\big(\mathrm{SR}(x_{\mathrm{LR}})&, x_{\mathrm{HR}}\big) +\\
   \lambda_{\mathrm{id}}\big(1-&\cos(\phi(\mathrm{SR}(x_{\mathrm{LR}})), \phi(x_{\mathrm{HR}}))\big)
\end{aligned}
\end{equation}
where $\ell_1$ is the pixel-wise $L_1$ distance, $\phi$ is the frozen backbone
embedding, and $\lambda_{\mathrm{id}}$ controls the impact of the identity constraint. \emph{Stage 2 (joint contrastive)}
warm-starts from Stage 1 and optimizes a contrastive
loss~\cite{hadsell2006contrastive} over $P{\times}K$-sampled HR/LR batches,
stepping the upsampler and PDT jointly while the backbone stays frozen.


\subsection{Baselines}
\paragraph{HR and synthetic LR backbones.} 
Our main reference point is the \emph{direct feed} baseline, in which the LR face is aligned and resized to the input layer size and then passed to an HR-trained backbone (EdgeFace-S or base) with \emph{no} particular front-end and no LR-specific training. It is the simplest deployment and, as we show, a hard baseline. We report it for both backbones so that every adaptation can be compared with the direct feed approach of the same model.

\begin{figure}[t]
  \centering
  \includegraphics[width=1.05\linewidth]{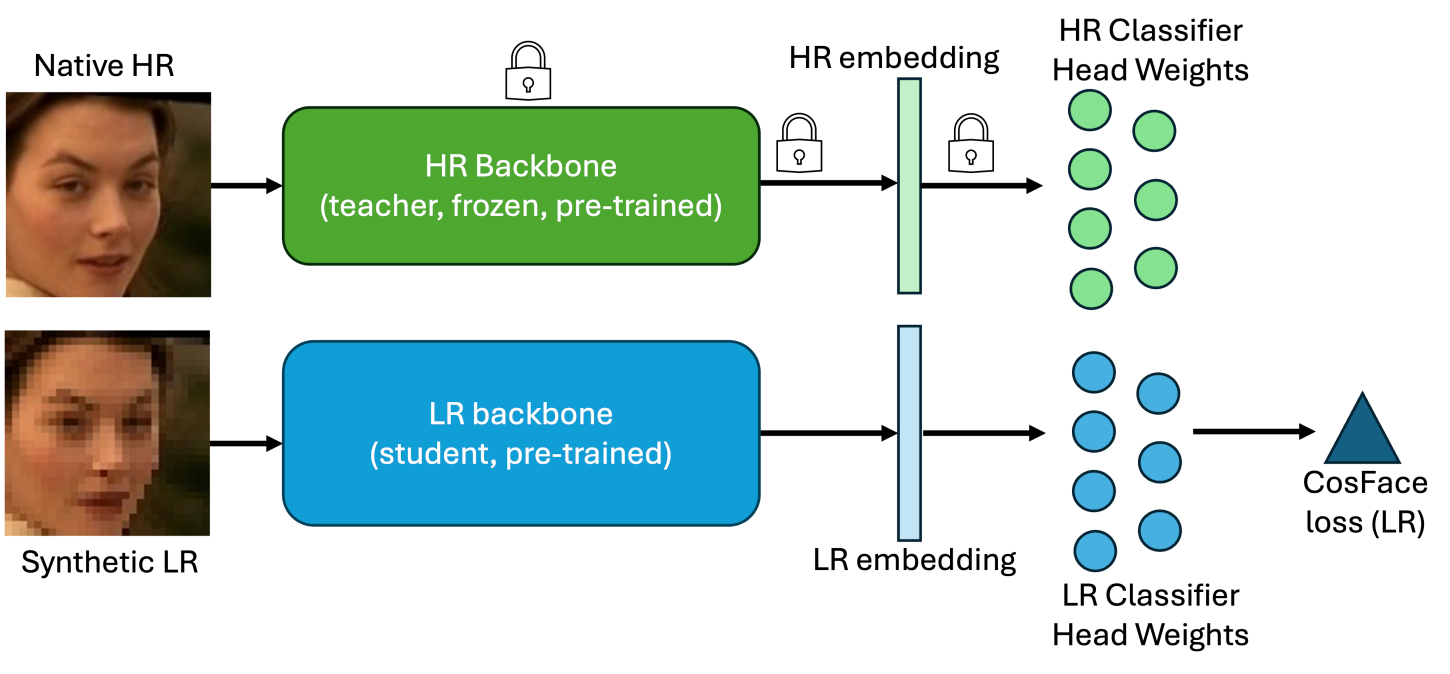}
  \caption{Adopted Knowledge Distillation (KD) paradigm using backbones trained on HR and LR data. The LR backbone is trained to produce embeddings similar to the HR version from LR data.}
  \label{fig:kd}
\end{figure}

\paragraph{Knowledge distillation (KD).}
A \emph{frozen} HR-trained EdgeFace-S teacher guides an LR student of identical
architecture (initialized from the LR backbone) via
$\mathcal{L}=\mathcal{L}_{\text{CosFace}}+\lambda_e\|\hat z_s-\hat z_t\|_2^2+\lambda_w\|\hat W_s-\hat W_t\|_2^2$,
with $\hat z$ the L2-normalized embeddings and $\hat W$ the L2-normalized sampled
PartialFC weights. We include KD as a retraining baseline (illustrated in
Fig.~\ref{fig:kd}).

\section{Experimental Setup}
\label{sec:setup}

\textbf{Training data.} WebFace4M~\cite{zhu2021webface260m}, in HR, the
interpolation LR variants (Sec.~\ref{sec:methods}), and the Real-ESRGAN $32$\,px
synthesis. \textbf{Backbones.} EdgeFace-S ($3.65$\,M) for compact models;
EdgeFace-base (frozen, WebFace12M) for the SR$+$PDT front-end.
\textbf{Synthetic benchmarks.} LFW~\cite{huang2008lfw},
CFP-FP~\cite{sengupta2016cfp}, AgeDB-30~\cite{moschoglou2017agedb} degraded to
$56/28/14/7$\,px in two face verification modalities: \emph{HR$\to$LR} (one image degraded, one HR)
and \emph{LR$\to$LR} (both degraded). We also report
standard HR verification and IJB-C~\cite{maze2018ijbc} TAR@FAR.
\textbf{Real native-LR.} TinyFace~\cite{cheng2018tinyface} identification (mAP,
Rank-$k$; gallery $\sim$157k) under two pipelines:  aligned crops with padding (\texttt{aligned\_pad})
and the Differentiable Face Aligner (DFA, ResNet-50) from CVLface~\cite{cvlface2024}.
All reported accuracies use horizontal-flip as an augmentation at test time and are given as percentages.

\section{The Synthetic--Real Gap}
\label{sec:gap}
We present our findings comparing baseline results on synthetic cross-resolution matching and synthetic training to real native LR evaluation sets.
\subsection{Synthetic cross-resolution}
Table~\ref{tab:macro} reports the macro-mean accuracy over LFW, CFP-FP and AgeDB-30 for the synthetic cross-resolution sets, by modality. Most LR-aware settings outperform the HR baseline, though not all: under HR$\to$LR the $28$\,\ac{} setting falls marginally below it for the LR backbone and for KD. The strongest single configuration is $28$\,\ca{}, and KD yields the largest gains. LR$\to$LR is consistently harder than HR$\to$LR at the same resolution: on the $28$\,\ca{} column the gap is largest for PDT, followed by the LR backbone and KD, and comes mainly from AgeDB-30 and CFP-FP. A single HR reference in the pair is an advantage that vanishes when both faces are degraded, which is one reason why synthetic HR$\to$LR results look optimistic. Table~\ref{tab:perbench} shows the same pattern per benchmark: under HR$\to$LR, $28$\,\ca{} is best on
\emph{every} benchmark for all three methods. Under LR$\to$LR (Table~\ref{tab:perbenchlr}) it still leads on aggregate, but the best setting per benchmark is less consistent (e.g.\ $56$\,\ca{} edges ahead on LFW).

\begin{table}[t]
\centering
\caption{Synthetic cross-resolution macro-mean accuracy (mean over LFW, CFP-FP,
AgeDB-30), by modality. Best LR-trained value per row in \textbf{bold}.
$28$\,\ca{} is best for every method under HR$\to$LR, and for all but PDT under
LR$\to$LR.}
\label{tab:macro}
\small
\setlength{\tabcolsep}{3.5pt}
\begin{tabular}{l c c c c c}
\toprule
Method & HR & 56\,\ca & 28\,\ac & 28\,\ca & 14\,\ac \\
\midrule
\multicolumn{6}{l}{\emph{HR\,$\to$\,LR}}\\
LR backbone & 78.85 & 80.25 & 78.77 & \best{80.74} & 79.67 \\
KD          & 78.85 & 80.74 & 78.60 & \best{82.19} & 79.82 \\
PDT         & 78.85 & 79.86 & 80.35 & \best{81.22} & 79.92 \\
\midrule
\multicolumn{6}{l}{\emph{LR\,$\to$\,LR}}\\
LR backbone & 78.28 & 78.92 & 78.74 & \best{79.06} & 78.31 \\
KD          & 78.28 & 79.14 & 78.91 & \best{80.59} & 78.51 \\
PDT         & 78.28 & 78.81 & \best{79.10} & 79.08 & 78.54 \\
\bottomrule
\end{tabular}
\end{table}

\begin{table}[t]
\centering
\caption{Per-benchmark synthetic HR$\to$LR mean accuracy. $28$\,\ca{} is
best for every synthetic benchmark and method (\textbf{bold}).}
\label{tab:perbench}
\begin{adjustbox}{width=1\linewidth,center}
\small
\setlength{\tabcolsep}{3.5pt}
\begin{tabular}{l l c c c c c}
\toprule
Bench. & Method & HR & 56\,\ca & 28\,\ac & 28\,\ca & 14\,\ac \\
\midrule
\multirow{3}{*}{LFW}    & LRb & 82.21 & 83.83 & 81.70 & \best{84.55} & 82.98 \\
                        & KD  & 82.21 & 84.36 & 81.90 & \best{86.39} & 83.43 \\
                        & PDT & 82.21 & 83.78 & 84.47 & \best{85.79} & 84.02 \\
\midrule
\multirow{3}{*}{CFP-FP} & LRb & 77.57 & 78.53 & 77.80 & \best{78.74} & 77.59 \\
                        & KD  & 77.57 & 79.09 & 77.82 & \best{80.87} & 77.59 \\
                        & PDT & 77.57 & 78.10 & 78.51 & \best{79.16} & 78.14 \\
\midrule
\multirow{3}{*}{AgeDB}  & LRb & 76.76 & 78.39 & 76.82 & \best{78.93} & 78.44 \\
                        & KD  & 76.76 & 78.76 & 76.07 & \best{79.32} & 78.44 \\
                        & PDT & 76.76 & 77.71 & 78.06 & \best{78.70} & 77.61 \\
\bottomrule
\end{tabular}
\end{adjustbox}
\end{table}

\begin{table}[t]
\centering
\caption{Per-benchmark synthetic \emph{LR$\to$LR} mean accuracy.
\textbf{Bold}: best LR-trained per row.}
\label{tab:perbenchlr}
\begin{adjustbox}{width=1\linewidth,center}
\small
\setlength{\tabcolsep}{3.5pt}
\begin{tabular}{l l c c c c c}
\toprule
Bench. & Method & HR & 56\,\ca & 28\,\ac & 28\,\ca & 14\,\ac \\
\midrule
\multirow{3}{*}{LFW}    & LRb & 84.83 & \best{85.21} & 84.88 & 85.06 & 84.23 \\
                        & KD  & 84.83 & 85.20 & 85.47 & \best{86.34} & 84.83 \\
                        & PDT & 84.83 & 85.13 & \best{85.56} & 85.45 & 85.54 \\
\midrule
\multirow{3}{*}{CFP-FP} & LRb & 77.41 & 77.89 & 77.41 & \best{78.26} & 77.53 \\
                        & KD  & 77.41 & 78.04 & 77.57 & \best{80.05} & 77.53 \\
                        & PDT & 77.41 & 78.09 & \best{78.41} & 78.25 & 77.84 \\
\midrule
\multirow{3}{*}{AgeDB}  & LRb & 72.60 & 73.66 & \best{73.93} & 73.86 & 73.17 \\
                        & KD  & 72.60 & 74.19 & 73.70 & \best{75.38} & 73.17 \\
                        & PDT & 72.60 & 73.22 & 73.34 & \best{73.54} & 72.25 \\
\bottomrule
\end{tabular}
\end{adjustbox}
\end{table}


\subsection{Distillation gives the largest synthetic gains}
Among the three adaptations, knowledge distillation yields the largest gains on HR$\to$synthetic LR (Table~\ref{tab:macro}). The student sees only the degraded image, but its target embedding is computed by the teacher from the pristine one, so the HR-shaped geometry supplies supervision that the LR input alone cannot carry. The effect is largest at the most extreme degradations ($14$ and $7$\,px), where the student has least signal of its own. These gains are on \emph{synthetic} low resolution, however, and do not carry over to real native LR.

\subsection{Real native-LR}
On real LR data the ranking changes. Table~\ref{tab:gap} compares the two evaluations of the \emph{same} compact models: synthetic HR$\to$LR macro-mean and real native-LR TinyFace mAP (DFA alignment). The setting that is best on synthetic LR, $28$\,\ca{}, is the worst of the family on real LR, below even the HR-trained model. On the other hand, the milder $56$\,\ca{} setting performs the best. Training at $56$\,px leaves the embeddings closer to the real native-LR domain than training at a resolution tuned on synthetic data.

\begin{table}[t]
\centering
\caption{The same EdgeFace-S models, \emph{synthetic} vs \emph{real} LR.
Synthetic: HR$\to$LR macro-mean Acc. Real: TinyFace mAP, $\times100$ (DFA). Best
LR-trained per row in \textbf{bold}; the resolution differs.}
\label{tab:gap}
\small
\setlength{\tabcolsep}{4pt}
\begin{tabular}{l c c c c}
\toprule
Evaluation & HR-tr. & 56\,\ca & 28\,\ca & 14\,\ac \\
\midrule
Synthetic macro (Acc) & 78.85 & 80.25 & \best{80.74} & 79.67 \\
Real TinyFace (mAP)   & 52.55 & \best{54.68} & 49.31 & --- \\
\bottomrule
\end{tabular}
\end{table}

\begin{figure}[t]
  \centering
  \includegraphics[width=0.8\linewidth]{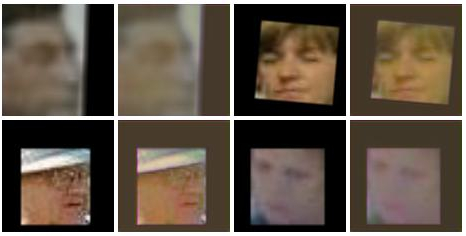}
  \caption{PDT translator outputs on native-LR TinyFace probes. Trained on synthetic LR, the translation appears to have a smoothing effect on the input images.}
  \label{fig:pdtgrid}
\end{figure}

The same lesson applies to input-side adaptation under \emph{both} alignments.
Table~\ref{tab:tfpad} reports TinyFace under the \texttt{aligned\_pad} crops: every PDT variant sits \emph{below} the HR baseline at every resolution, because a translator fit to synthetic LR does not transfer to genuine degradation and mostly smooths the probe (Fig.~\ref{fig:pdtgrid}). Likewise, KD trails the plain LR backbone. 
Both observations motivate us to test a more realistic
degradation model and a learned way to invert it, evaluated on real LR.

\begin{table}[t]
\centering
\caption{Real native-LR TinyFace under \texttt{aligned\_pad} crops (mAP / R-1,
$\times100$). Best per row in \textbf{bold}. 
}
\label{tab:tfpad}
\begin{adjustbox}{width=1\linewidth,center}
\small
\setlength{\tabcolsep}{3.5pt}
\begin{tabular}{l c c c c c}
\toprule
Method & HR & 56\,\ca & 28\,\ac & 28\,\ca & 14\,\ac \\
\midrule
LR backbone (mAP) & 47.24 & \best{49.31} & 49.23 & 48.51 & 45.67 \\
KD (mAP)          & 47.24 & 48.96 & 47.46 & 47.94 & 45.67 \\
PDT (mAP)         & 47.24 & 46.69 & 46.19 & 45.22 & 43.66 \\
\midrule
LR backbone (R-1) & 54.48 & 56.14 & \best{56.41} & 55.69 & 52.28 \\
PDT (R-1)         & 54.48 & 53.78 & 53.49 & 52.31 & 51.10 \\
\bottomrule
\end{tabular}
\end{adjustbox}
\end{table}

\section{Generative Pipeline: Results}
\label{sec:results}
We show our results on medium to high synthesis efforts tested on real-world LR images.

\subsection{Native-LR identification}
Table~\ref{tab:tinyface} is our result on real native-LR TinyFace (DFA
alignment), with every model compared against the direct feed of its own
backbone. Generative methods are in \emph{italic}.

\begin{table}[t]
\centering
\caption{Real native-LR \textbf{TinyFace} identification (DFA alignment).
``base''$=$frozen EdgeFace-base (WebFace12M); ``S''$=$EdgeFace-S ($3.65$\,M);
\emph{direct} feeds the aligned LR face with no front-end. Generative methods in
\emph{italic}. \textbf{Bold}$=$best compact (S) model; \underline{underline}$=$best
short of the large direct-feed reference. All values $\times100$.}
\label{tab:tinyface}
\begin{adjustbox}{width=1\linewidth,center}
\small
\setlength{\tabcolsep}{3.5pt}
\begin{tabular}{l c c c c c}
\toprule
Model & mAP & R-1 & R-5 & R-10 & R-20 \\
\midrule
base, direct (HR) & 65.12 & 71.08 & 74.97 & 75.83 & 76.98 \\
\emph{base $+$ RRDB-SR $+$ PDT} & \underline{57.53} & 64.70 & 70.52 & 72.26 & 74.20 \\
\emph{base $+$ ESPCN-SR $+$ PDT} & 56.64 & 63.95 & 69.74 & 71.62 & 73.44 \\
\midrule
S, HR-trained (direct) & 52.55 & 59.66 & 66.63 & 69.05 & 71.49 \\
S, $56$\,\ca{} aug & \best{54.68} & \best{61.40} & 68.16 & 70.28 & 72.80 \\
\emph{S, Real-ESRGAN $32$ stem} & 53.74 & \best{61.40} & 68.62 & 70.84 & 73.47 \\
S, $56$\,\ca{} PDT & 51.94 & 58.88 & 66.17 & 68.51 & 70.79 \\
S, $28$\,\ca{} aug & 49.31 & 57.16 & 64.14 & 66.71 & 69.21 \\
\bottomrule
\end{tabular}
\end{adjustbox}
\end{table}

On the compact backbone, which has no strong direct feed to fall back on, simple
LR-aware augmentation is the only synthesis that helps. A mild $56$\,\ca{} training
gives the best compact model, ahead of both its own direct feed and the
Real-ESRGAN $32$\,px stem. This is the regime our study targets and the
one place where synthesis pays off.

\begin{figure}[t]
  \centering
  \begin{subfigure}{1\linewidth}\includegraphics[width=\linewidth]{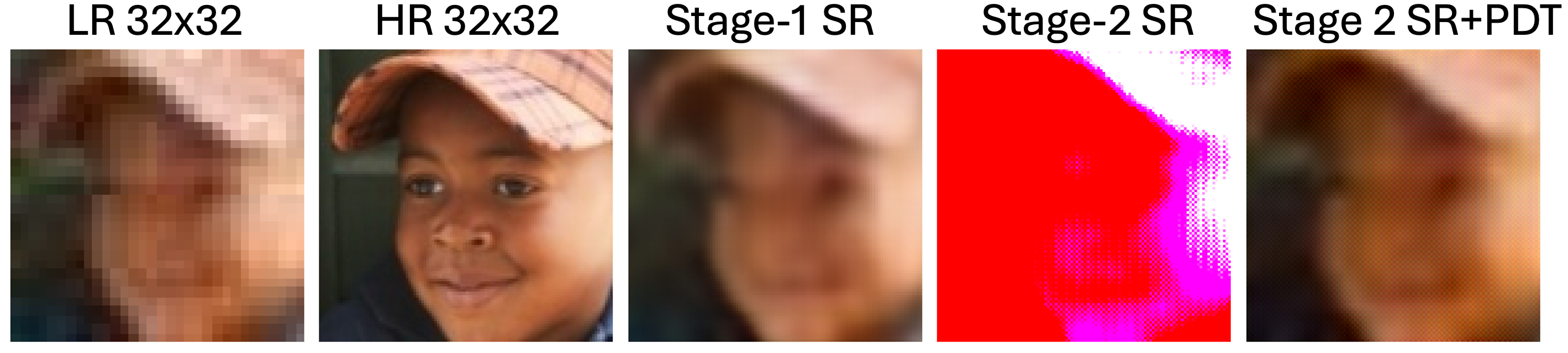}\caption{ESPCN sub-pixel}\end{subfigure}\hfill
  \begin{subfigure}{1\linewidth}\includegraphics[width=\linewidth]{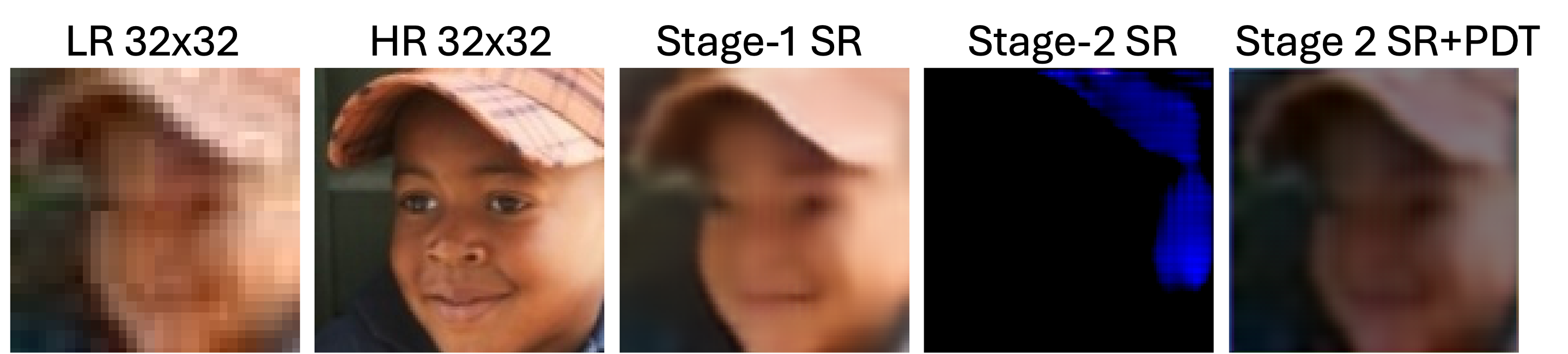}\caption{RRDB (Real-ESRGAN)}\end{subfigure}
  \caption{Learned $32\!\to\!112$ reconstructions on WebFace4M pairs after
  identity-aware pretraining (input / SR output / HR target).}
  \label{fig:srviz}
\end{figure}

In contrast, the learned generative front-end does not help as much. On the frozen
EdgeFace-base the full SR$+$PDT pipeline trails a direct feed of the aligned LR
image, and the SR stage on its own is worse still (Table~\ref{tab:compactsr}).
Learned upscaling therefore loses to the interpolation already performed by the aligner resulting in lowered front-end accuracy. The PDT
alone shows the same pattern on the compact backbone, again below its direct
feed. Fig.~\ref{fig:srviz} shows the upsampler does recover plausible and
identity-consistent detail, but this visual quality does not translate into
recognition gains.

\subsection{Comparison with the literature on TinyFace}
Table~\ref{tab:sota} compares our compact models with the efficient face
recognizers of the EFaR~2023 competition~\cite{kolf2023efar}, which reports
TinyFace identification for sub-$5$\,M-parameter models under a common protocol.
All our EdgeFace-S models are trained \emph{from scratch for $100$ epochs}, the
budget used for our LR backbones, whereas the competition's
EdgeFace-S($\gamma{=}0.5$) is the fully-trained official checkpoint of the same
architecture. Even on this shorter schedule our HR-trained EdgeFace-S lands close
to the official checkpoint, which indicates that our protocol and alignment are
comparable, unlike cross-paper numbers collected under
different alignment, which can differ by tens of points.

At an equal parameter budget and the same schedule, LR-aware augmentation outperforms the official checkpoint of EdgeFace-S on both Rank-1 and Rank-5. Other methods, such as Modified-MobileFaceNet, are also more accurate at a cost of $1.5\times$ the FLOPs. The larger gains come not from the backbone but from training and adaptation: the learned SR front-end on a stronger frozen backbone reaches the level of the best compact competitor (Table~\ref{tab:tinyface}), while a direct feed of that same backbone is far ahead of everything in Table~\ref{tab:sota}. At comparable complexity, native-LR accuracy is driven by LR-aware synthesis and adaptation rather than by
architecture or a longer training schedule.

\begin{table}[t]
\centering
\caption{TinyFace identification (\%) against efficient models from the EFaR~2023
competition~\cite{kolf2023efar} (their Table~4). All models are compact
($\leq$$4.1$\,M parameters); the training-schedule difference is discussed in the
text. \textbf{Bold}: best at the $3.65$\,M budget.}
\label{tab:sota}
\small
\setlength{\tabcolsep}{4pt}
\begin{tabular}{l c c c c}
\toprule
Method & Params & MFLOPs & R-1 & R-5 \\
\midrule
\multicolumn{5}{l}{\emph{EFaR~2023 competition~\cite{kolf2023efar}}}\\
SAM-MFaceNet               & 1.10M & 236.8 & 61.31 & 66.33 \\
EdgeFace-XS($\gamma$=0.6)  & 1.77M & 154.0 & 58.77 & 63.89 \\
Mod.\ MobileFaceNet        & 2.10M & 456.9 & 64.29 & 69.60 \\
EdgeFace-XS-Q              & 2.24M & 196.9 & 59.20 & 64.75 \\
EdgeFace-S($\gamma$=0.5)   & 3.65M & 306.1 & 61.02 & 65.47 \\
GhostFaceNetV1-1           & 4.09M & 215.7 & 59.54 & 64.75 \\
\midrule
\multicolumn{5}{l}{\emph{Ours (same EdgeFace-S architecture)}}\\
EdgeFace-S, HR-trained     & 3.65M & 306.1 & 59.66 & 66.63 \\
EdgeFace-S, $56$\,\ca{} aug & 3.65M & 306.1 & \best{61.40} & \best{68.16} \\
\bottomrule
\end{tabular}
\end{table}

\subsection{LR training as a regularizer (HR benchmarks)}
\begin{figure}[t]
  \centering
  \includegraphics[width=1\linewidth]{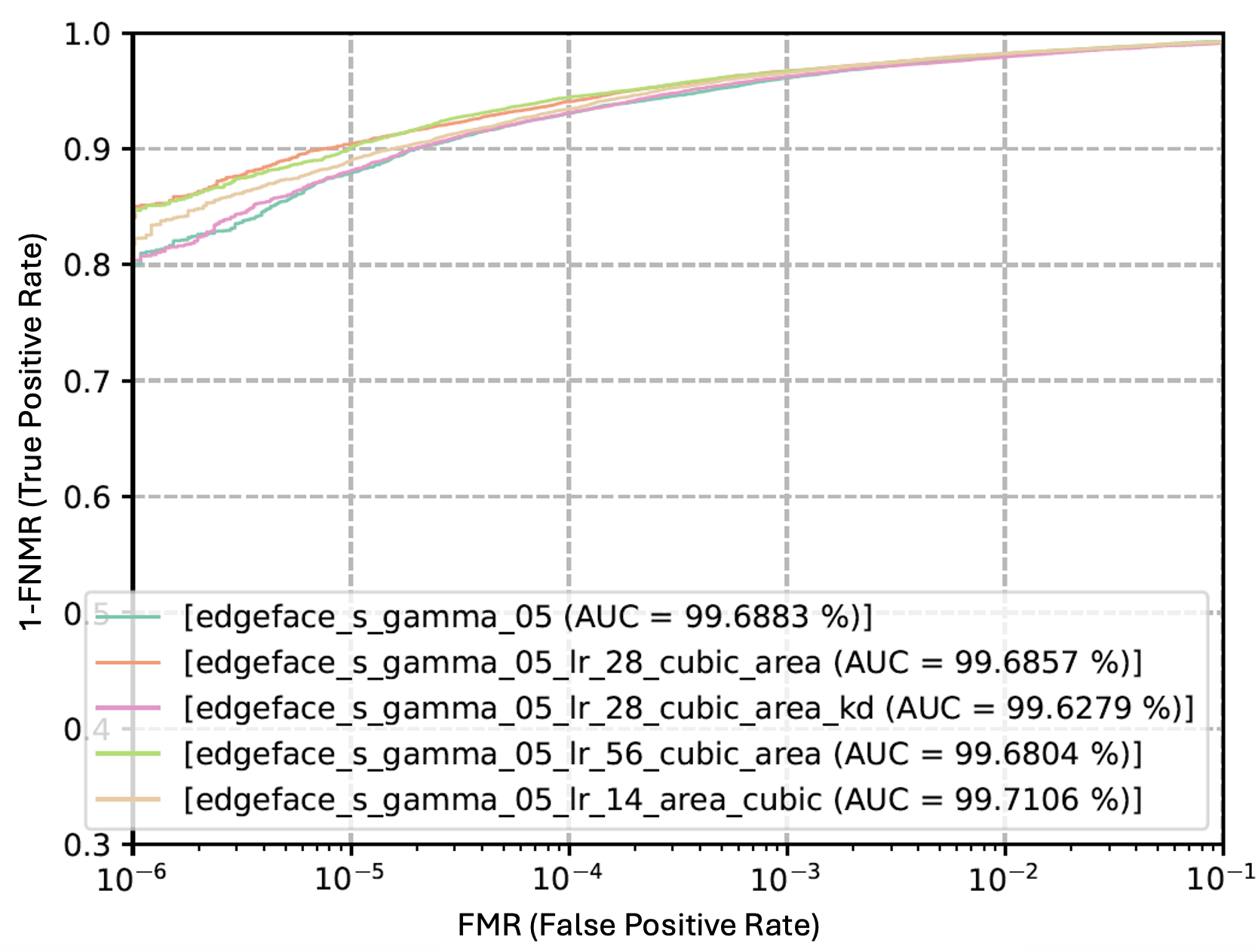}
  \caption{IJB-C False Match Rate (FMR) against $1-$False Non-Match Rate (FNMR).
  The LR-trained EdgeFace-S variants match or exceed the HR baseline across
  operating points.}
  \label{fig:ijbc}
\end{figure}

A useful side effect (Table~\ref{tab:hr}): training on LR-augmented data at $56$
and $28$\,px \emph{matches or exceeds} the HR baseline on every standard
benchmark and on IJB-C (Fig.~\ref{fig:ijbc}). The more aggressive $14$\,\ac{}
setting is the exception, falling below the baseline on five of the six
verification sets while still improving model performance on IJB-C. The $28$\,\ca{} backbone is best
or tied on four of the six verification sets, and $28$\,\ac{} gives the best
IJB-C TAR@FAR. The native $32$\,px stem, trained only on very-low-resolution input size performs worse on HR:  specializing for native LR trades away HR FR capability, showing to be suitable only when the input is expected to present LR artifacts.
We also found that Distillation largely maintains and outperforms different tests against the HR baseline on most sets, although the $28$\,px KD variant performs worse overall and mostly on CFP-FP and AgeDB-30. 

\begin{table}[t]
\centering
\caption{Standard HR verification (Acc\,$\times100$) for the KD models.
Distillation mostly preserves the LR-training regularization; \textbf{bold} marks
the best per row.}
\label{tab:hrkd}
\small
\setlength{\tabcolsep}{3pt}
\begin{tabular}{l c c c c c}
\toprule
 & HR & 56\,\ca & 28\,\ac & 28\,\ca & 14\,\ac \\
\midrule
LFW       & 99.45 & \best{99.62} & 99.48 & 99.45 & 99.58 \\
CALFW     & 94.53 & \best{95.12} & 94.50 & 94.33 & 94.77 \\
CPLFW     & 90.90 & \best{91.37} & 90.70 & 90.85 & 91.18 \\
CFP-FF    & 99.29 & \best{99.47} & 99.13 & 99.03 & 99.39 \\
CFP-FP    & 96.53 & 97.06 & 95.63 & 95.81 & \best{97.11} \\
AgeDB-30  & 94.38 & \best{95.72} & 93.88 & 93.10 & 94.98 \\
\bottomrule
\end{tabular}
\end{table}

\begin{table}[t]
\centering
\caption{Standard HR verification (Acc\,$\times100$) and IJB-C TAR@FAR\,(\%) for the
LR backbones; the last column is the native $32$\,px stem (\emph{n$32$}), which collapses on HR.}
\label{tab:hr}
\begin{adjustbox}{width=1\linewidth,center}
\small
\setlength{\tabcolsep}{3pt}
\begin{tabular}{l c c c c c c}
\toprule
 & HR & 56\,\ca & 28\,\ac & 28\,\ca & 14\,\ac & \emph{n$32$} \\
\midrule
LFW       & 99.45 & 99.58 & 99.53 & \best{99.65} & 99.58 & 93.48 \\
CALFW     & 94.53 & 95.25 & 95.28 & \best{95.32} & 94.52 & --- \\
CPLFW     & 90.90 & 91.63 & \best{91.85} & \best{91.85} & 89.97 & --- \\
CFP-FF    & 99.29 & \best{99.57} & \best{99.57} & 99.46 & 99.03 & --- \\
CFP-FP    & 96.53 & \best{97.24} & 97.10 & 97.09 & 96.11 & 81.01 \\
AgeDB-30  & 94.38 & 95.63 & 95.63 & \best{95.67} & 93.90 & 75.21 \\
IJB-C\,$10^{-3}$ & 96.11 & 96.68 & \best{96.91} & 96.71 & 96.51 & --- \\
IJB-C\,$10^{-4}$ & 93.07 & 94.44 & \best{94.56} & 94.06 & 93.43 & --- \\
\bottomrule
\end{tabular}
\end{adjustbox}

\end{table}

\subsection{Impact of the learned SR front-end}
To isolate the contribution of the learned super-resolution front-end, we evaluate the identity-aware SR (ESPCN) pretraining checkpoint in isolation before the joint contrastive stage with the PDT (Table~\ref{tab:compactsr}). On both backbones the SR pretraining alone underperforms a direct feed of the aligned LR image. Adding the joint contrastive stage recovers much of this loss on EdgeFace-base but still does not reach the direct feed. This reinforces our central finding: a learned SR module is not useful as a standalone front-end, and even with joint training, the added effort does not beat feeding a strong backbone the aligned image.

\begin{table}[t]
\centering
\caption{Effect of the learned SR front-end on native-LR TinyFace (DFA, \%).
``SR only'' is the identity-aware SR (ESPCN) pretraining checkpoint without the
joint PDT stage; ``$+$PDT'' adds the joint contrastive stage.}
\label{tab:compactsr}
\small
\setlength{\tabcolsep}{4pt}
\begin{tabular}{l c c c}
\toprule
Model & mAP & R-1 & R-5 \\
\midrule
EdgeFace-S, direct feed      & 52.55 & 59.66 & 66.63 \\
EdgeFace-S $+$ SR only       & 35.82 & 44.02 & 53.19 \\
\midrule
EdgeFace-base, direct feed   & 65.12 & 71.08 & 74.97 \\
EdgeFace-base $+$ SR only    & 43.48 & 52.23 & 60.76 \\
EdgeFace-base $+$ SR $+$ PDT & 56.64 & 63.95 & 69.74 \\
\bottomrule
\end{tabular}
\end{table}

\section{Ablations and Analysis}
\label{sec:ablation}
\textbf{Interpolation pair.} Cubic downsampling preserves more high-frequency structure, which is then smoothed by area upsampling. The cubic-down/area-up pair ($28$\,\ca) is the strongest synthetic configuration, while area-down/cubic-up ($28$\,\ac) trails it on most HR$\to$LR sets yet leads on IJB-C. When training, only $4$ of the $12$ down/up combinations converged stably. This shows that the interpolation choice directly affects optimization.

\textbf{Resolution variation.} Accuracy degrades smoothly with training resolution on synthetic LR but the \emph{best} training resolution depends on the target domain: $28$\,px on synthetic, $56$\,px on native LR, with $14$\,px below the HR baseline on TinyFace. At $7$\,px all methods degrade sharply on CFP-FP and AgeDB-30, an unresolved domain gap.

\begin{table}[t]
\centering
\caption{Adaptation strategies by synthesis effort and accuracy on TinyFace (DFA aligned). We show different EdgeFace variants (model), the synthesis degradation type. ``Real degr.'' indicates whether the synthesis
reproduces genuine sensor/optics degradation rather than naive interpolation.}
\label{tab:effort}
\begin{adjustbox}{width=1\linewidth,center}
\small
\setlength{\tabcolsep}{3pt}
\begin{tabular}{l l l c c}
\toprule
Strategy & Effort & Model & Real degr. & mAP\,(\%) \\
\midrule
\textbf{Interp.\ aug ($56$\,\ca)}      & \textbf{low}    & \textbf{S}       & \textbf{no} & \textbf{54.68} \\
Interp.\ aug ($28$\,\ca)      & low    & S        & no  & 49.31 \\
Real-ESRGAN $32$ stem         & medium & S        & yes & 53.74 \\
\midrule
SR$+$PDT (ESPCN)      & high   & base$^\dagger$ & yes & 56.64 \\
SR$+$PDT (RRDB)       & high   & base$^\dagger$ & yes & 57.53 \\
\midrule
Direct feed (no synthesis)    & none   & S & --  & 52.55 \\
\bottomrule
\end{tabular}
\end{adjustbox}
\\[2pt]
{\footnotesize $^\dagger$frozen EdgeFace-base; others retrain EdgeFace-S.}
\end{table}

\textbf{Does synthesis effort pay off?} Not monotonically
(Table~\ref{tab:effort}). On the compact backbone the \emph{lowest}-effort
option, a single $56$\,px interpolation, beats both the no-synthesis baseline and
the medium-effort degradation stem while $28$\,px---the setting that was optimal on
synthetic data---is the worst among all. The \emph{highest}-effort option is
the best \emph{adaptation} in the table, but only on the large frozen backbone,
and it is still beaten by feeding that backbone directly. Effort therefore pays
off only where it changes the probe embedding most: cheap augmentation for a
retrainable compact model, and a learned front-end only when the backbone is
fixed and strong.

\textbf{Where capacity matters.} A strong frozen backbone already absorbs much
degradation: direct feed of EdgeFace-base beats every adaptation, including our
learned SR front-end on the same backbone, so a trainable front-end is worth most
when the backbone is small or cannot be used directly. The two backbone groups in
Tables~\ref{tab:tinyface} and~\ref{tab:effort} differ in capacity, training
sets (WebFace4M vs.\ WebFace12M) and whether they are frozen, therefore the accuracy is only comparable within each backbone model variant (S and base). 
It also explains why KD wins
synthetic extremes but not native LR: anchoring to an HR-oriented embedding helps when the test degradation is synthetic and mild, and hurts when the target domain
genuinely differs.

\textbf{Where does the gap come from?} Two factors could explain the poor
transfer to TinyFace: a \emph{degradation} mismatch (interpolation does not
reproduce the blur, noise, and compression of real optics) and an
\emph{alignment} mismatch (landmarks are far less reliable on genuinely captured
LR faces). Our results implicate both but separate them. The synthetic optimum
($28$\,px) being the worst configuration on real LR (Table~\ref{tab:gap}) points
at the degradation model, since all compared models share the same alignment
pipeline. Alignment instead confounds absolute accuracy. Moving from
\texttt{aligned\_pad} crops to the DFA aligner raises every model substantially,
so any TinyFace comparison must fix the aligner. We therefore attribute the gap
primarily to the degradation model, while noting that detection and alignment on
native LR remain a major source of absolute error.

\subsection{Bias and fairness evaluation}
\begin{table}[t]
\centering
\caption{Per-group FMR, its ratio $r$ to the geometric mean, and the Gini index
of per-group FMR on RFW at TMR@FMR$=10^{-2}$.}
\label{tab:fairness}
\begin{adjustbox}{width=\linewidth}
\small
\setlength{\tabcolsep}{4pt}
\begin{tabular}{l l cc cc cc cc c}
\toprule
 & & \multicolumn{2}{c}{African} & \multicolumn{2}{c}{Asian} & \multicolumn{2}{c}{Caucasian} & \multicolumn{2}{c}{Indian} & \\
\cmidrule(lr){3-4}\cmidrule(lr){5-6}\cmidrule(lr){7-8}\cmidrule(lr){9-10}
px & Model & FMR & $r$ & FMR & $r$ & FMR & $r$ & FMR & $r$ & Gini \\
\midrule
\multirow{3}{*}{HR} & HR-tr. & 0.74 & \textbf{2.80} & 0.27 & 1.01 & 0.07 & 0.25 & 0.37 & 1.39 & 0.37 \\
 & 56\,\ca & 1.10 & \textbf{3.36} & 0.53 & 1.62 & 0.03 & 0.10 & 0.60 & 1.82 & 0.36 \\
 & 28\,\ca & 1.37 & \textbf{4.24} & 0.60 & 1.85 & 0.00 & 0.05 & 0.80 & 2.47 & 0.39 \\
\midrule
\multirow{3}{*}{56} & HR-tr. & 0.77 & \textbf{3.41} & 0.43 & 1.92 & 0.03 & 0.15 & 0.23 & 1.03 & 0.41 \\
 & 56\,\ca & 0.90 & \textbf{3.25} & 0.57 & 2.04 & 0.00 & 0.06 & 0.70 & 2.51 & 0.33 \\
 & 28\,\ca & 1.31 & \textbf{3.85} & 0.43 & 1.28 & 0.03 & 0.10 & 0.70 & 2.06 & 0.41 \\
\midrule
\multirow{3}{*}{28} & HR-tr. & 0.33 & \textbf{3.36} & 0.13 & 1.34 & 0.00 & 0.17 & 0.13 & 1.33 & 0.42 \\
 & 56\,\ca & 0.60 & \textbf{3.92} & 0.33 & 2.17 & 0.00 & 0.11 & 0.17 & 1.08 & 0.45 \\
 & 28\,\ca & 0.64 & \textbf{2.59} & 0.43 & 1.76 & 0.03 & 0.14 & 0.40 & 1.62 & 0.31 \\
\bottomrule
\end{tabular}
\end{adjustbox}
\end{table}
\label{sec:fairness}
We measure the demographic fairness on RFW~\cite{wang2019rfw}, which contains African, Asian, Caucasian and Indian subjects. We evaluate the three EdgeFace-S backbones at their respectively calibrated thresholds on IJB-C at TMR@FMR$=10^{-2}$, using the same IJB-C protocol as in Table~\ref{tab:hr} on RFW. The probes in RFW are further degraded with bicubic downsampling and upsampling to $56/28/14/7$\,px depending on the model.
Table~\ref{tab:fairness} reports, for each group $d$, its FMR as a percentage and the ratio $r_d=\mathrm{FMR}_d/g$, where $g$ is the geometric mean of the FMR over the four groups, so $r_d=1$ when the group's FMR equals the geometric mean. The last column is the Gini coefficient of the FMR over the four groups, which is $0$ when the four groups have equal FMR and grows with the difference between them. In the table the px column is the test resolution and the model name is the training regime. A group with no false matches has its FMR set to $0.5/3000$, half the smallest non-zero rate.

The Caucasian group has better verification accuracy than the other three groups: its $r$ stays far below $1$ in every cell, while the African $r$ is the highest throughout, with the Asian and Indian groups in between. This ordering, with the Caucasian group most accurate and the African group least accurate, agrees with bias previously reported on RFW~\cite{wang2019rfw}.

Training on low-resolution data does not systematically reduce this difference. The $56$\,\ca{} and $28$\,\ca{} backbones, which give the highest average accuracy in Tables~\ref{tab:macro} and~\ref{tab:tinyface}, show an FMR Gini that moves in both directions relative to the HR-trained model depending on the test resolution, with no consistent improvement in either backbone. The largest group ratio anywhere in the table also belongs to the $28$\,\ca{} model. Accuracy gains from LR-aware synthesis therefore do not carry over into reduced demographic disparity.

We report resolutions of $28$\,px and above. At $14$ and $7$\,px the global FNMR is above $0.9$ for all models, so the fixed threshold no longer separates the mated and non-mated scores and the per-group comparison is not informative.

\section{Limitations and Future Work}
\label{sec:limits}
Our learned SR front-end is evaluated only on a large frozen backbone, where a direct feed is already a hard baseline. We leave the more informative setup, a \emph{compact} backbone that cannot exploit such a feed, as future work. All our LR training data are synthetic by design as we study the setting in which \emph{no} labelled native-LR identities are available. We therefore do not evaluate the common practice of pre-training on synthetic LR and fine-tuning on limited real LR. We leave for future work using native LR images, such as TinyFace's training split, for training synthesis strategies and assessing LR FR performance. 
Quantifying how much real LR data each strategy needs before it becomes worthwhile is the clearest next step, alongside learning the \emph{paired} HR-reference/LR-probe conditional directly (e.g.\ a conditional or diffusion generator, evaluated on real probes against the direct-feed baseline) and analyzing the fairness of LR capture and sampling.
\section{Conclusion}
We studied five adaptation strategies for low-resolution face recognition under scarce native-LR data and evaluated all of them on real LR. Our four conclusions are as follows: \emph{(i)}~The degradation setting that is optimal on synthetic benchmarks is not the one that is optimal on real LR: $28$\,\ca{} is best across the synthetic benchmarks yet is the worst configuration on TinyFace, where the milder $56$\,px setting wins. \emph{(ii)}~Synthesis effort does not pay off monotonically; on a retrainable compact backbone the cheapest interpolation augmentation beats both Real-ESRGAN-style degradation and a learned SR front-end. \emph{(iii)}~A learned, identity-aware super-resolution front-end never beats simply feeding a strong frozen backbone the aligned LR image, so a direct-feed baseline should be reported before any restoration or translation pipeline is claimed to help. \emph{(iv)}~The average-accuracy gains from LR-aware synthesis do not reduce demographic bias: the FMR disparity on RFW shows no consistent improvement. In practice we therefore recommend matching synthesis effort to the deployment constraint, and validating on real native LR rather than on synthetic degradation alone.

{\small
\section*{Acknowledgments}
We acknowledge the funding provided by the following projects: Frontex under the
Frontex Research Grants Programme, Call 2024/CFP/INNOVATE/01, Grant Agreement
No.\ 2025/280; CarMen project, HORIZON-CL3-2023-BM-01, no.\ 101168325; PopEye
project, HORIZON-CL3-2023-BM-01, no.\ 101168317; and the CERTAIN Project, HORIZON-CL4-2024-DATA-01, no.\ 101189650. 
}

{\small
\bibliographystyle{ieee}
\bibliography{references}
}

\end{document}